\documentclass{article}

\usepackage{iclr2026_conference,times}
\iclrfinalcopy

\usepackage{amsmath}
\usepackage{amssymb}
\usepackage{booktabs}
\usepackage[ruled,linesnumbered]{algorithm2e}
\usepackage{graphicx}

\usepackage{tikz}
\usetikzlibrary{positioning,shapes.geometric,arrows.meta,fit,backgrounds,calc}
\usepackage{pgfplots}
\pgfplotsset{compat=1.18}

\usepackage[protrusion=true,expansion=false]{microtype}

\usepackage{xspace}

\usepackage{xurl}
\usepackage{hyperref}

\usepackage{xcolor}

\newcommand{\model}{SkillEvolver\xspace}
\newcommand{\sccontrol}{SkillCreator-SkillsBench\xspace}

\title{\model: Skill Learning as a Meta-Skill}

\author{%
  Genrui Zhang\textsuperscript{2}\thanks{Equal contribution.} \quad
  Erle Zhu\textsuperscript{1}\footnotemark[1] \quad
  Jinfeng Zhou\textsuperscript{1} \quad
  Caiyan Jia\textsuperscript{2} \quad
  Hongning Wang\textsuperscript{1}\thanks{Correspondence to: \texttt{hw-ai@tsinghua.edu.cn}.} \\[0.6ex]
  \textsuperscript{1}Tsinghua University \quad
  \textsuperscript{2}Beijing Jiaotong University
}

\begin{document}
\maketitle

% Override the ICLR final-copy running header. This is an arXiv preprint, not
% an ICLR acceptance, so we replace "Published as a conference paper at ICLR
% 2026" with "Preprint.".
\lhead{Preprint.}

\begin{abstract}
Agent skills today are \emph{static artifacts}: authored once -- by human curation or one-shot generation from parametric knowledge -- and then consumed unchanged, with no mechanism to improve from real use. We propose \textbf{\model}, a lightweight, plug-and-play solution for \emph{online skill learning}, in which a single \emph{meta-skill} iteratively authors, deploys, and refines \emph{domain-specific skills}. The learning target of \model is the skill's prose and code, not model weights, so that the resulting artifact drops into any agent without retraining; and the meta-skill itself is just another skill, loaded through the same interface by any protocol-compliant CLI-agent. Unlike trace-distillation, the meta-skill refines only \emph{after} deploying the learnt skill, such that the learning signal comes from failures another agent encounters while using it -- not from exploratory traces alone. Refinement iterations are governed by a fresh-agent overfit audit that catches possible leakage as well as deployed-skill-specific failures, including the silent-bypass mode in which a skill appears valid in content but is never invoked at runtime. On $83$ SkillsBench tasks spanning $15^{+}$ domains, \model reaches $56.8\%$ accuracy versus $43.6\%$ for curated human skills and $29.9\%$ for the no-skill baseline; on three GPU kernel optimization tasks from KernelBench, it also raises mean speedup from $1.16$ to $1.51$ on average.
\end{abstract}

\section{Introduction}
\label{sec:intro}

Modern LLM agents increasingly require \emph{procedural} knowledge to tackle complex real-world problems: not merely what a task is about, but how it should be carried out in a specific environment. For example, how to fill a domain-specific spreadsheet template, follow a project-specific schema, invoke a brittle tool interface, or avoid known failure modes in a recurring workflow. Agent skills have emerged as a lightweight mechanism for encoding such knowledge. Specifically, a skill is a short, task-specific artifact loaded at inference time, typically bundling natural-language instructions, executable scripts, reference files, examples, and usage constraints into a reusable dependency for an agent~\citep{anthropic2025agentskills, anthropic2025claudecode,xuyan2026agentskills}. Unlike one-off prompts or demonstrations, skills can be stored, transferred, revised and redeployed, thus turning local procedural know-how into a portable unit of agent behavior. Human-curated skills have been shown to substantially improve agents' performance on skill-focused benchmarks, such as SkillsBench~\citep{skillsbench2026}, and ecosystem-scale studies report hundreds of thousands of community-authored skills in circulation~\citep{li2026ecosystem}. Yet this skill-authoring paradigm still relies heavily on human expertise. In long-tail deployment settings, where new domain tasks arise on demand, recruiting a specialist to author for every workflow is neither timely nor economical. This raises the central question: \textit{can an agent acquire procedural knowledge from a bounded set of deployment-time trials and package it as a reusable skill, without retraining model weights?}

Recent efforts to automate skill creation mainly fall into two directions, but neither assume deployment settings of on-demand, few-trial skill authoring regime  we target. \emph{Parametric self-generation} -- the model writes a skill directly from its pre-trained knowledge, as in Anthropic's official skill-creator~\citep{anthropic2025skillcreator} and the self-generation condition of SkillsBench~\citep{skillsbench2026} -- commits without grounded feedback. And its one-shot variant can perform no better than, and sometimes worse than, using no skill at all~\citep{skillsbench2026}. 
\emph{Trace- and RL-based skill acquisition} (RL, Reinforcement Learning), by contrast, derives its strength from broad execution coverage: {Trace2Skill}~\citep{trace2skill2026} mines roughly $200$ trajectories per domain through a multi-stage trace-distillation pipeline over a pre-collected trajectory pool; 
{SkillRL}~\citep{skillrl2026} grows a recursively expanding skill library by pooling experiences across many training tasks; 
and a broader family of RL-based acquisition methods build on similar cross-task aggregation~\citep{xuyan2026agentskills}. These solutions are powerful, but they assume substantial offline preparation per domain --- a pre-collected trajectory pool, a multi-stage distillation pipeline, or a cross-task RL loop. Many real-world tasks arrive on demand, one at a time, with only a handful of exploration trials affordable before a skill must ship. Our focus is therefore not ecosystem-scale management or large-pool offline consolidation~\citep{li2026ecosystem}, but \emph{on-demand domain-specific skill learning}: how to author, deploy, audit, and refine a reusable skill within the bounded experience of a single newly-arrived task.

In this work, we propose \textbf{\model}, a lightweight solution framework for \emph{online skill learning}: adapting an external skill artifact for a newly arrived task, rather than updating model parameters. This mirrors the timing of test-time adaptation~\citep{sun2020testtime,wang2021tent}, but \model updates a reusable skill by observing a small number of training-time task trials and redeploying the revised artifact to fresh downstream agents. \model uses a single \emph{meta-skill} to drive a standard command-line interface (CLI)-agent through the lifecycle of skill acquisition: exploring a small number of trials on the task's available training split, authoring a reusable \emph{domain skill}, deploying the candidate skill to fresh Domain-Skill Agents, and refining the artifact from their observed behavior (\S\ref{sec:method-overview}). What distinguishes our approach from prior skill-generation and trace-distillation work is that refinement is grounded in this deployment handoff rather than in the authoring agent's self-reflection: a candidate can fail not only by producing the wrong answer, but also by omitting a needed instruction, exposing a misleading procedure, or being silently bypassed by the using agent. A fresh-session Auditor then gates each synthesized candidate for leakage, overfitting, and deployment-specific failures before it can enter the accepted skill sequence. This design turns a bounded set of training-time trials on a newly arrived task into a reusable procedural artifact for future agents.

Our contributions are summarized as follows:
\begin{itemize}
\item We formulate online skill learning as the acquisition of reusable procedural artifacts, and introduce \model, a plug-and-play framework (meta-skill) that lets a standard CLI-agent author domain-specific skills without retraining model weights (\S\ref{sec:method}).
\item We introduce a deployment-grounded refinement loop that tests each candidate skill as actual dependency used by fresh agents. Combined with strategy-diversified trials and an independent Auditor, \model exposes skill failures that are not visible from exploratory traces alone (\S\ref{sec:method-sampling}--\ref{sec:method-audit}).
\item We evaluate \model on SkillsBench~\citep{skillsbench2026}, covering $83$ tasks across $15^{+}$ domains under a uniform train/test split. \model reaches $\mathbf{56.8\%}$ accuracy (avg@$5$), against $43.6\%$ for curated human skills and $29.9\%$ for the no-skill baseline (\S\ref{sec:exp-main}). On three GPU kernel optimization tasks from KernelBench~\citep{ouyang2025kernelbenchllmswriteefficient}, it also raises mean speedup from $1.16$ to $\mathbf{1.51}$ on average, suggesting that the method transfers from binary workflow benchmarks to continuous-optimization scenarios. Ablation analysis shows that iterative refinement accounts for a substantial part of the SkillsBench gain (\S\ref{sec:exp-ablation}), and additional analyses show that evolved skills also reduce the agent's token usage, interaction length, and wall-clock time at validation (\S\ref{sec:exp-efficiency}).
\end{itemize}

\section{Related Work}
\label{sec:related}

\paragraph{Automated skill and context creation.} \textbf{SkillsBench}~\citep{skillsbench2026} establishes the $A$/$B$/$C$ framing we build on ($A$: no skill, $B$: curated human skill, $C$: self-generated skill) and reports that blind parametric-knowledge-only skill generation hurts agent's performance. \textbf{ACE}~\citep{ace2025}, building on the adaptive memory of Dynamic Cheatsheet~\citep{suzgun2025dynamic}, evolves an in-context playbook via iterative edits rather than parameter updates. Our refinement pass (\S\ref{sec:method-synthesize}) shares that incremental-edit idea but produces a self-contained domain skill that a different agent can later load. Earlier agent scaffolds -- ReAct~\citep{yao2022react}, Reflexion~\citep{shinn2023reflexion}, Self-Refine~\citep{madaan2023selfrefine}, Voyager~\citep{wang2023voyager} -- and persistent-memory systems such as MemGPT~\citep{packer2023memgpt} and generative agents~\citep{park2023generative} equip an agent at inference time; but they do not author a transferable skill as a separate deliverable.

\paragraph{Trace distillation and RL-based skill acquisition.} \textbf{Trace2Skill}~\citep{trace2skill2026} establishes trace distillation as a viable primitive for agent skill creation, mining around 200 trajectories per domain through a hand-engineered Python pipeline (per-trace patch proposal, hierarchical inductive merge) on three domain pools (spreadsheet, VisionQA, math reasoning). \textbf{SkillRL}~\citep{skillrl2026} instead grows a recursively expanding skill library by pooling experiences across many training tasks under an RL loop, and a broader family of RL-based skill acquisition methods follow similar cross-task aggregation~\citep{xuyan2026agentskills}. 
Both lines derive their strength from breadth of coverage --- a per-domain trajectory pool or a cross-task training distribution --- whereas \model targets a single new task with only a few deployment-time trials (typically four), without any per-domain pipeline and shared instance pool. We cite these as prior work that motivated our direction rather than as baselines: comparing a pool-based or cross-task method to a task-level method on a task-level benchmark would not be fair in either direction.

\paragraph{Oracle access and evaluation discipline.} Reading training-label files -- test specifications, reference solutions -- is legitimate in any supervised-learning framework; but the problem is whether the test-set oracle leaks into the artifact evaluated at test time. Contamination concerns in LLM evaluation have become a recurring problem~\citep{sainz2023contamination}; the agent setting adds a new surface, since the authoring agent itself may read training-label files and encode them into the skill. We address this with a strict train/test split (the curated training skill is deleted at source, so it is never reachable) and a workspace whitelist enforced via a PreToolUse hook (Appendix~\ref{app:anticheat}).

\paragraph{Broader agent paradigms and benchmarks.} Skills sit within a broader line of work on agent scaffolding, including tool-use models~\citep{schick2023toolformer}, code-as-action loops~\citep{wang2024codeact}, and multi-agent orchestration~\citep{hong2023metagpt, wu2023autogen}, and they are evaluated alongside a wider wave of agent benchmarks spanning code, web, OS, and enterprise settings~\citep{jimenez2024swebench, zhou2024webarena, liu2024agentbench, xie2024osworld, xu2024theagentcompany, mialon2024gaia}. Our focus is orthogonal to these lines: \model is a method for producing the skill artifacts these agents can leverage, not a new agent scaffold or benchmark suite.

\section{Method}
\label{sec:method}

\label{sec:method-overview}
\begin{figure*}[t]
  \centering
  \makebox[\textwidth][c]{\includegraphics[page=1,trim=30.8 234.1 24.6 8.0,clip,width=.9\textwidth]{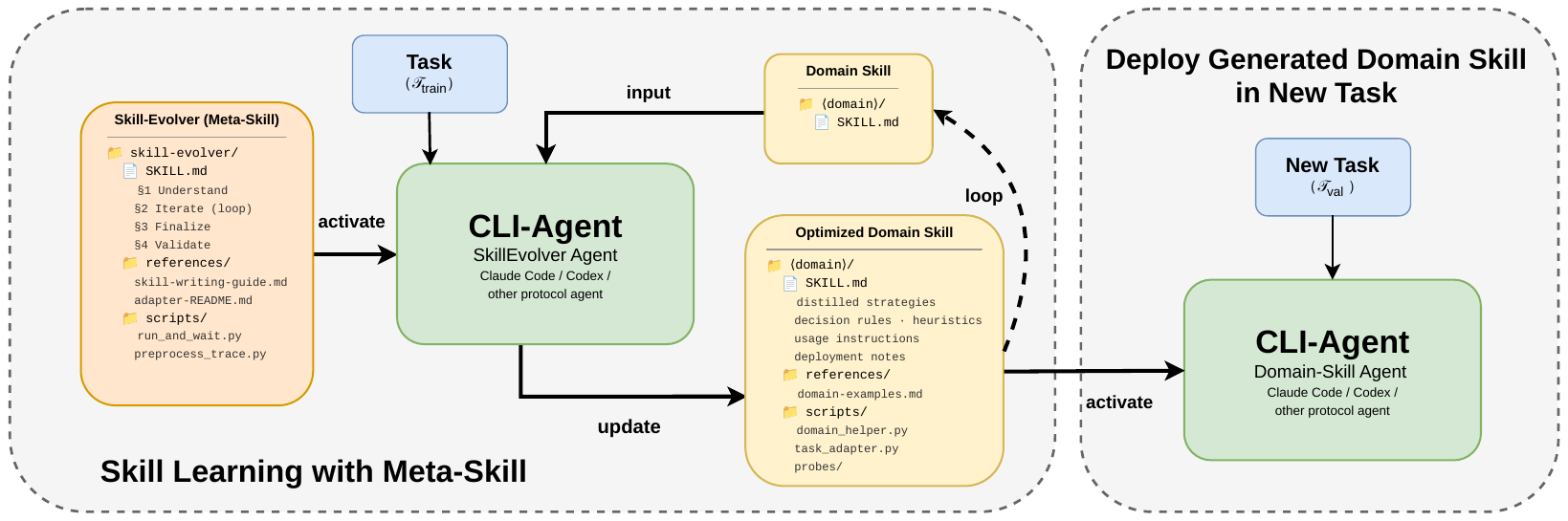}}
\caption{\textbf{\model\ as a portable meta-skill.} \model is a meta-skill that any CLI-agent that loads skills (Claude Code, Codex, \ldots) can load through the same interface used for any domain skill. Given a new task $\mathcal{T}{=}(\mathcal{T}_{\text{train}},\mathcal{T}_{\text{val}})$ with a held-out validation split, the CLI-agent uses the meta-skill to iteratively construct, test, and update a deployment-ready domain skill $v^{*}$. The learned object is itself a portable skill, containing prose instructions, scripts, references, and examples rather than updated model weights. Figure~\ref{fig:pipeline} details the internal learning loop.}
  \label{fig:deployment}
\end{figure*}

Most methods for improving an LLM agent have it learn from trial-and-error~\citep{shinn2023reflexion}. At each iteration $r$, the agent attempts the training task $K$ times, producing trajectories $\{\tau_{r,i}\}_{i=1}^{K}$ with task rewards $y_{r,i}$~\citep{song2024eto}. 
These trajectories are then analyzed to extract lessons, which methods like Trace2Skill distill into a skill artifact, namly a domain-specific skill, that the agent loads without parameter updates~\citep{trace2skill2026}.
\model{} follows this overall pattern with the domain skill itself as the update target. Figure~\ref{fig:deployment} shows the deployment surface, Figure~\ref{fig:pipeline} unpacks one iteration, and we provide pseudocode in Appendix~\ref{app:evolver-alg}.

% ------------------------------------------------------------------
% Figure 1 — Deployment surface: SkillEvolver as a portable meta-skill
% ------------------------------------------------------------------

% ------------------------------------------------------------------
% Figure 2 — One iteration of the SkillEvolver loop (full width)
% ------------------------------------------------------------------
\begin{figure*}[t]
  \centering
  \includegraphics[page=2,trim=15.7 357.0 15.7 23.8,clip,width=0.9\textwidth]{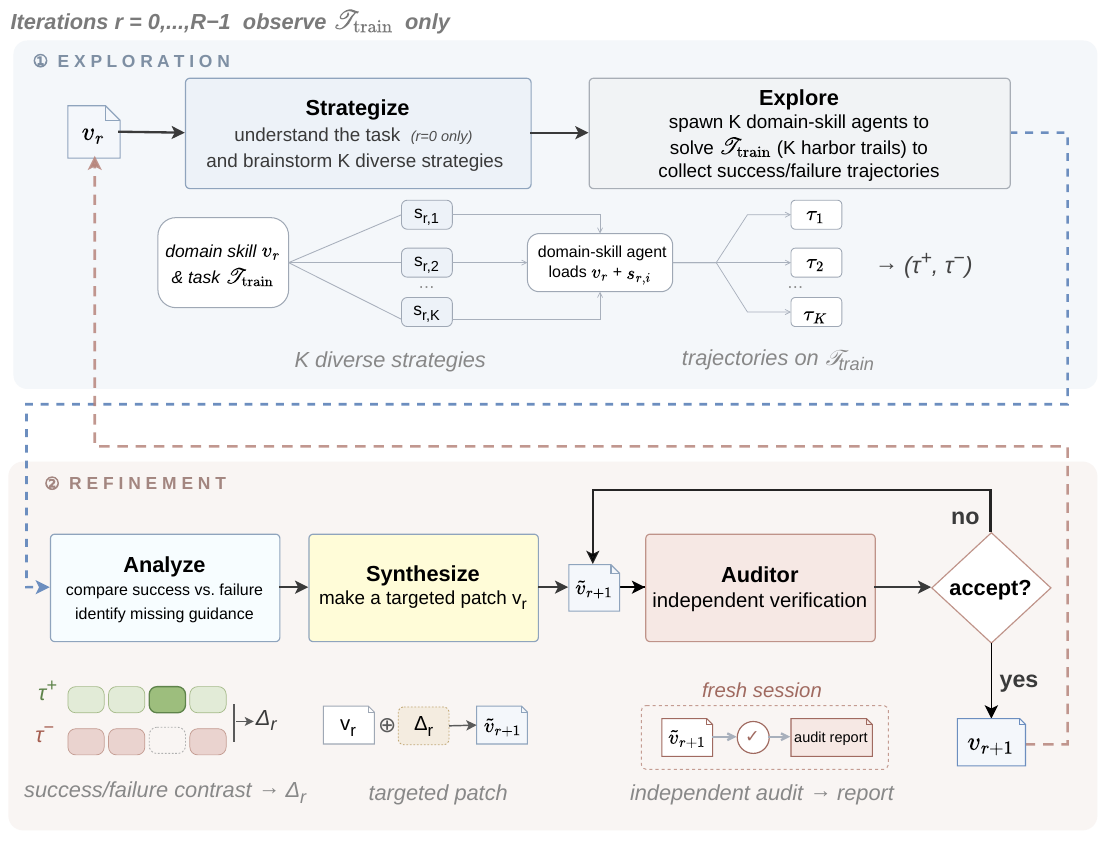}
\caption{\textbf{One iteration of the \model loop.} At iterations $r = 0, \ldots, R-1$, \model observes only $\mathcal{T}_{\text{train}}$. Starting from the current skill $v_r$, the agent explores $K$ training-time trials, analyzes success and failure traces, synthesizes a targeted revision $v_{r+1}$, and audits it in an independent fresh session. Approved revisions continue through the loop; failed audits trigger another targeted patch. After the loop, \model finalizes a deployable skill before the held-out split $\mathcal{T}_{\text{val}}$ is used for evaluation.
The whole process is executed by a \model\ Agent (a CLI-agent equipped with the \model\ skill plugin).
}
  \label{fig:pipeline}
\end{figure*}

\subsection{\model\ as a meta-skill}
\label{sec:method-metaskill}

% ----- citation anchor: hu2024adas (ADAS, ICLR 2025, arXiv:2408.08435) -----
% Verbatim source for "meta agent":
%   Abstract: "...new agents can be automatically discovered by a meta
%   agent programming ever better ones in code."
%   Section 1 (first body mention, with scare quotes):
%   "...a 'meta' agent programming ever better ones in code."
% What we borrow: the meta-X naming pattern.
% What we differ on: in ADAS the agent itself is meta (it programs new
%   agents in code); in our setting the agent is held fixed and the SKILL.md
%   artifact plays the meta role. We make this difference explicit so readers
%   familiar with ADAS do not import the wrong intuition.
\textbf{\model} is a \emph{meta-skill}: Rather than solving a target task directly, it
instructs a CLI-agent, i.e., the \emph{\model\ Agent}, to author, refine, and deploy a reusable domain skill for that task. We adopt the \emph{meta-X} naming convention following \citet{hu2024adas}, who introduce a \emph{meta agent} as a foundation model that programs new agents in code. Our setting differs in what plays the meta role. We hold the agent fixed: it can be any CLI-agent that can load skills, such as Claude Code or Codex. The meta role is instead taken by the \emph{skill}, which instructs the agent to author another skill rather than modify itself.

The learning signal is not the \model\ Agent's reflection on its own execution trajectories, but what a \emph{separate} CLI-agent (given only the candidate domain skill and the task) actually does when handed that skill; we call this second agent the \emph{Domain-Skill Agent}. In each iteration, %(a SkillEvolver loop), 
the \model\ Agent deploys the current candidate skill, spawns multiple Domain-Skill Agents to attempt the task with it, and refines the skill from what the Domain-Skill Agents did or failed to do.

\subsection{The \model\ loop}
\label{sec:method-iteration}

An iteration of the \model\ loop takes a candidate skill $v_r$ and a training task $\mathcal{T}_{\text{train}}$, and returns a refined candidate $v_{r+1}$, where each candidate is a domain skill --- a directory of prose and scripts loadable by any CLI-agent. The refined candidate is obtained by applying a patch to $v_r$, where the patch is derived from a contrast over labelled trajectories produced by a Domain-Skill Agent running $v_r$ on $\mathcal{T}_{\text{train}}$.

\subsubsection{Strategy-diversified exploration}
\label{sec:method-sampling}

We call our sampling procedure \emph{strategy-diversified exploration}.
Before running $K$ parallel rollouts at iteration $r$, the \model\ Agent writes a strategy set $\mathcal{S}_r=\{s_{r,i}\}_{i=1}^{K}$, where each $s_{r,i}$ specifies a distinct high-level solution plan, and assigns one fresh Domain-Skill Agent to each strategy.  Unlike diversity from token-level sampling, this treats diversity as coverage over high-level decision axes such as library choice, algorithm family, and instruction interpretation. The resulting rollout is
\begin{equation}
\label{eq:explore}
(\tau_{r,i}, y_{r,i}) =
\mathrm{Trial}(v_r, \mathcal{T}_{\text{train}}, s_{r,i}), \qquad i=1,\ldots,K ,
\end{equation}
where $\tau_{r,i}$ is the trajectory, $y_{r,i}$ is the reward, and $s_{r,i}$ is the strategy assigned to the $i$-th rollout at iteration $r$. We use $K{=}4$ throughout the work.

The strategies are not sampled by raising model temperature. Temperature changes local wording and tool-call details, but the resulting agents often share the same high-level plan. Instead, the \model\ Agent writes explicit strategy files before launch and checks that no two strategies are identical on all major axes. A second check marks each concrete training constant as either invariant or parametric; for every parametric axis, at least one strategy must derive the value at runtime rather than copy the training value.

At the first iteration ($r{=}0$), no domain skill has been learned yet. We therefore deploy a minimal skill that only assigns rollout $i$ to its strategy file $s_{0,i}$ before the Domain-Skill Agent attempts the task. Thus, even the bootstrap rollouts are controlled by explicit strategy files rather than left as free-form attempts.

At later iterations ($r{>}0$), $v_r$ is already a domain-specific skill, so we place the strategy assignment before the rest of the skill text and update the strategy files to target failure modes observed in the previous iteration.

\subsubsection{Contrastive skill update}
\label{sec:method-synthesize}

% ----- citation anchor: song2024eto (ETO, ACL 2024, arXiv:2403.02502) -----
% Verbatim source for the success/failure trajectory pairing construct:
%   Abstract: "...gathering failure trajectories to create contrastive
%   trajectory pairs."
%   Abstract: "...the agent utilizes these trajectory preference pairs to
%   update its policy using contrastive learning methods like DPO."
%   Section 3.2 (formal): "we get the contrastive trajectory dataset
%   D_p = {(u, e_w, e_l)^(i)}|D_p|_{i=1}."
% Note: the words "successful trajectory" / "failed trajectory" are ambient
% RL vocabulary (no single paper coined them), but ETO is the first to pair
% success+failure trajectories as a contrastive dataset for LLM agents.
% That pairing is what we credit, not the words themselves.
Instead of using labeled trajectories to update the agent policy, we use them to update the deployable skill artifact. Given trajectories $\{(\tau_{r,i}, y_{r,i})\}_{i=1}^K$ with task rewards $y_{r,i}$, the \model\ Agent compares high-reward and low-reward trials to identify what the current skill is missing. For binary-reward tasks such as SkillsBench, these sets are the passing and failing trials; for scalar-reward tasks such as KernelBench, they are the top- and bottom-scoring trials:
\[
\tau_r^{+} = \mathrm{Top}(\{\tau_{r,i}\}_{i=1}^K; y_{r,i}), \qquad
\tau_r^{-} = \mathrm{Bottom}(\{\tau_{r,i}\}_{i=1}^K; y_{r,i}).
\]
The contrastive signal is then summarized as
\begin{equation}
\label{eq:gap}
\Delta_r = \phi(\tau_r^{+}) \setminus \phi(\tau_r^{-}),
\end{equation}
where $\phi$ denotes an LLM-based reading function that extracts task-relevant features from a set of trajectories rather than a programmatic parser. At $r{=}0$, the contrast asks what winners knew that losers lacked. At $r{>}0$, where $v_r$ is already deployed, it instead asks where the skill misled, under-specified, or failed to guide the Domain-Skill Agent.

% ----- citation anchors: shinn2023reflexion + trace2skill2026 -----
% shinn2023reflexion (Reflexion, NeurIPS 2023, arXiv:2303.11366):
%   Section 1: "We propose Reflexion, a new paradigm for 'verbal'
%   reinforcement that parameterizes a policy as an agent's memory
%   encoding paired with a choice of LLM parameters."
%   Section 1: "Reflexion has several advantages compared to more
%   traditional RL approaches like policy or value-based learning:
%   1) it is lightweight and doesn't require finetuning the LLM..."
%   --> source of the "verbal reinforcement" framing.
% trace2skill2026 (Trace2Skill, arXiv:2603.25158):
%   Section 5: "we strictly study frozen-model, training-free,
%   artifact-level adaptation, ensuring our distilled skills remain
%   entirely model-agnostic."
%   Section 2: "construct an improved skill from trajectories on
%   D_evolve, without updating theta."
%   --> source of "artifact-level adaptation"; closest published
%   analogue to our setup.
% Alternative anchor for the gradient-free framing: TextGrad's
% "textual gradients" (Yuksekgonul 2024, arXiv:2406.07496). Swap in
% if a reviewer prefers a more algorithmic flavor.
Where Exploration-Based Trajectory Optimization (ETO)~\citep{song2024eto} uses such pairs as preference data to update an LLM agent's policy via DPO~\citep{rafailov2024directpreferenceoptimizationlanguage}, we use the contrast as \emph{verbal reinforcement}~\citep{shinn2023reflexion} to refine a frozen-weight agent's skill document --- an instance of artifact-level adaptation~\citep{trace2skill2026}. In compact form, the synthesis step is

\begin{equation}
\label{eq:patch}
\tilde{v}_{r+1} = \mathrm{Patch}(v_r, \Delta_r).
\end{equation}

The patch is natural-language and code content written into the skill artifact rather than into model parameters; no weights are touched at any iteration. The main filter is whether a candidate feature would likely be known from pretraining alone. If so, it is not added.

Synthesis applies $\Delta_r$ as a localized edit to the skill artifact. At $r{=}0$, the edit creates the first domain skill $v_1$ from the contrastive signal and any reusable code observed in high-reward traces. At $r{>}0$, it patches $v_r$ rather than rewriting it, preserving working guidance while adding only the missing constraint, code pattern, or tool exposed by the latest contrast. When executable scripts are included, they must operate on inputs supplied at runtime rather than on filenames, constants, or answers copied from the training instance.

\subsubsection{Independent audit and finalization}
\label{sec:method-audit}

After synthesis, \model\ invokes an Auditor: a separate CLI-agent session with a clean context, used to verify the candidate skill before it can be accepted into the loop. The Auditor receives only the candidate skill, the task instruction, the training data, and the labelled traces $\{(\tau_{r,i}, y_{r,i})\}$, not validation data or the \model\ Agent's context. As an artifact-level verifier, it checks whether the candidate is self-contained, grounded in observed traces, abstracted away from training-instance constants, and structured so that a fresh Domain-Skill Agent can apply it without relying on the evolver's private reasoning.

The audit returns a binary gate and a set of named violations,
\begin{equation}
\label{eq:audit}
(a_r, E_r) = \mathrm{Audit}(\tilde{v}_{r+1}, \mathcal{T}_{\text{train}}, \{(\tau_{r,i}, y_{r,i})\}_{i=1}^{K}).
\end{equation}
The checks are listed in Table~\ref{tab:auditor}. They cover both standard overfitting risks (instance-borrowed framing, hardcoded literals, and untraceable claims) and deployment-specific risks, such as whether strategy axes are abstracted, whether a primary script is surfaced where a Domain-Skill Agent will read it, and whether the skill can be silently bypassed.

A clean audit promotes the candidate to the next accepted skill, $v_{r+1}=\tilde{v}_{r+1}$, whereas an audit failure converts the reported violation into the next refinement target. The loop terminates when the accepted skill has no audited defect and the latest exploration traces expose no actionable failure mode, or when the iteration budget is exhausted. Before validation, \model\ selects from the accepted skills $\{v_j\}_{j=1}^{r+1}$, optionally applies a small hand merge, and uses training pass rate, trace cost, and generalization risk to choose the artifact written to the deployment directory for validation on $\mathcal{T}_{\text{val}}$.

% Table 1 (Auditor checks) moved to Appendix~\ref{app:auditor}.
% Anti-cheating subsection moved to Appendix~\ref{app:anticheat}.
% Algorithm 1 moved to Appendix~\ref{app:evolver-alg}.

\section{Experiments}
\label{sec:experiments}

In this section, we evaluate \model\ on SkillsBench~\citep{skillsbench2026} and %on
three GPU kernel optimization tasks from KernelBench~\citep{ouyang2025kernelbenchllmswriteefficient}. 
We first compare self-evolved skills with no-skill, human-curated, and self-generated baselines (\S\ref{sec:exp-main}). We then ablate the refinement loop by contrasting the one-pass pipeline ($R{=}1$) with the full two-iteration pipeline ($R{=}2$) (\S\ref{sec:exp-ablation}), analyze the cost--quality trade-off and downstream agent efficiency (\S\ref{sec:exp-efficiency}), and break down the gains across the SkillsBench skill-utility taxonomy (\S\ref{sec:exp-analysis}). 
Appendix~\ref{app:cases} and~\ref{app:cases-kb} present representative case studies to explain the main success and failure modes.

\subsection{Experimental Setup}
\label{sec:exp-setup}

\textbf{Benchmarks.} We evaluate on two benchmarks. The first is SkillsBench~\citep{skillsbench2026}, $87$ tasks across $15^{+}$ professional domains; we use the $83$-task subset with complete published no-skill and human-curated baselines and run all trials under Harbor~\citep{harbor2026}. Exploration and refinement operate on the training variant $\mathcal{T}_{\text{train}}$; validation is held out on $\mathcal{T}_{\text{val}}$ (anti-cheating: Appendix~\ref{app:anticheat}). 
The second benchmark is KernelBench~\citep{ouyang2025kernelbenchllmswriteefficient}, where we evaluate on three GPU kernel optimization tasks: \texttt{deepnarrowmlp}, \texttt{shufflenet}, and \texttt{gru}. These tasks cover three distinct model families: a dense MLP, a lightweight CNN, and a recurrent model. SkillsBench reports held-out workflow success as Avg@$5$, whereas KernelBench scores each trial by a correctness-weighted speedup objective.

\textbf{Conditions and trials.} We compare six conditions on the same agent harness (Claude Opus $4.6$ + Claude Code~\citep{anthropic2025claudecode}): (1) \textit{No skill}; (2) \textit{Human-curated skill}; (3) \textit{Self-Gen}~\citep{skillsbench2026}, a single LLM call that emits a skill from the task instruction; (4) \textit{\sccontrol}, our subagent-based adaptation of Anthropic's official skill-creator~\citep{anthropic2025skillcreator,anthropic2025agentskills} in which every human touchpoint is replaced by an Eval-Designer / Grader / Analyzer subagent (Appendix~\ref{app:scsb}); (5)~\textit{\model} ($R{=}1$), the non-refining ablation; and (6)~\textit{\model} ($R{=}2$), the full pipeline. \sccontrol{} sees the same context as \model, namely the task instruction and training-task environment $\mathcal{T}_{\text{train}}$, and differs only in the authoring mechanism, so it is the closest matched baseline to \model. \model\ uses $K{=}4$ exploration trials per iteration and $V{=}5$ validation trials, for $2K{+}V{=}13$ Harbor trials per task at $R{=}2$ and $K{+}V{=}9$ at $R{=}1$; the control runs only the $V{=}5$ validation trials in Harbor (its authoring iterations are local subprocess sessions). Per-task caps are \$$15$ and $200$ turns. On the three KernelBench tasks we keep the same $K$, $V$, and $R$ settings but optimize directly on scalar reward rather than binary pass/fail. Due to resource constraints, we do not run non-SkillEvolver
baselines on KernelBench.

\textbf{Metrics.} Following the SkillsBench convention, the per-task score is avg@$V{=}n_{\text{pass}}/n_{\text{trials}}$ (so $4/5{=}0.8$, not binary), and the headline aggregate is the per-task mean across the $83$ paper-scope tasks. Each condition is evaluated as a single sweep with $V{=}5$ independent Harbor trials per task per condition; we report the mean and identify per-task wins, ties and losses against the curated baseline. For KernelBench, the metric is a five-trial mean speedup score under a fixed skill: given a skill, the domain-skill agent runs $V{=}5$ independent Harbor trials, producing five candidate kernels, and we average their scalar rewards. For a single trial, the reward is defined as $\text{correctness}\in\{0,1\}\times\text{speedup}$, where $\text{speedup}=\frac{t_{\text{PyTorch}}}{t_{\text{kernel}}}$ is measured on an NVIDIA H100 against the reference PyTorch~\citep{paszke2019pytorch} implementation. Thus an incorrect kernel receives zero reward, while a correct kernel receives its measured runtime speedup. Unlike SkillsBench, the relevant quantity is therefore not pass rate but the average correctness-weighted speedup achieved over five runs.

\subsection{Skill Quality Comparison}
\label{sec:exp-main}

Table~\ref{tab:main} reports the headline comparison. On SkillsBench, \model\ at $R{=}2$ reaches avg@$5$ mean of $\mathbf{56.87\%}$, exceeding the no-skill baseline ($29.9\%$) by $+27.0$\,percentage points and the human-curated skill ($43.6\%$) by $+13.3$\,percentage points. The non-refining ablation, \model\ at $R{=}1$, already reaches $48.2\%$ ($+4.6$\,percentage points over the human-curated skill). The parametric self-generation baseline reported by~\citet{skillsbench2026} (Self-Gen, $32.0\%$) sits within a few points of the no-skill baseline. \sccontrol{} --- our subagent-based adaptation of Anthropic's skill-creator and the most informative baseline because it is given the same training-task context as \model\ and differs only in its authoring mechanism --- reaches $33.9\%$, also within a few points of the no-skill baseline and well below the human-curated skill. \model\ ($R{=}2$) outperforms every prior self-generated baseline by $\geq\!22$\,percentage points. 
On the three GPU kernel optimization tasks from KernelBench, \model\ also improves the main benchmark metric. At $R{=}2$, mean speedup increases on all three tasks, from $1.027$ to $1.089$ on \texttt{deepnarrowmlp}, from $1.117$ to $1.218$ on \texttt{shufflenet}, and from $1.326$ to $2.226$ on \texttt{gru}, corresponding to an average increase from $1.16$ to $1.51$. The largest gain appears on \texttt{gru}, where the evolved skill adds nearly $+0.9$ absolute reward over the no-skill baseline, suggesting that the learned artifact can capture nontrivial optimization knowledge even in a recurrent-model setting. The smaller but still positive gains on \texttt{deepnarrowmlp} and \texttt{shufflenet} indicate that the same authoring loop can also produce incremental optimization wins in MLP and CNN settings rather than only in a single favorable architecture family. At $R{=}1$, two of the three tasks already improve, but the gains are less stable than on SkillsBench.

\begin{table*}[t]
\centering
\footnotesize
\setlength{\tabcolsep}{1.4pt}
\renewcommand{\arraystretch}{1.16}
\caption{Main results across two evaluation settings. SkillsBench columns report avg@$5$ on the $83$-task paper scope, bucketed by the SkillsBench domain taxonomy; each cell shows the metric on the first line and the gain over \emph{No skill} on the second. The \emph{Overall} column reproduces the headline $83$-task aggregate; per-domain columns are computed over each domain's tasks with measured pipeline runs (per-task coverage visualized in Figure~\ref{fig:appendix-heatmap}). KernelBench columns report the mean speedup score averaged over $V{=}5$ validation runs on three representative continuous-reward optimization tasks, again with the gain over \emph{No skill} shown below the metric. \emph{Other} aggregates the four smallest SkillsBench groups: manufacturing, energy, mathematics, and health.}
\label{tab:main}
\resizebox{\textwidth}{!}{%
\begin{tabular}{@{}l c c c c c c c c c c c c@{}}
\toprule
\textbf{} & \multicolumn{9}{c}{\textbf{SkillsBench (avg@$5$)}} & \multicolumn{3}{c}{\textbf{KernelBench (Speedup)}} \\
\cmidrule(lr){2-10}\cmidrule(l){11-13}
\textbf{Method} & \textbf{Overall} & \textbf{SW Eng} & \textbf{Office} & \textbf{Sci.} & \textbf{Media} & \textbf{Fin.} & \textbf{Cyber} & \textbf{Robot.} & \textbf{Other} & \textbf{D.\ MLP} & \textbf{ShuffleNet} & \textbf{GRU} \\
 & {\tiny $n{=}83$} & {\tiny $n{=}16$} & {\tiny $n{=}14$} & {\tiny $n{=}12$} & {\tiny $n{=}11$} & {\tiny $n{=}8$} & {\tiny $n{=}8$} & {\tiny $n{=}5$} & {\tiny $n{=}9$} & {\tiny $n{=}1$} & {\tiny $n{=}1$} & {\tiny $n{=}1$} \\
\midrule
\textit{No skill}               & $29.9$ & $35.0$ & $24.3$ & $38.3$ & $29.1$ & $15.0$ & $27.5$ & $32.0$ & $40.0$ & $1.027$ & $1.117$ & $1.326$ \\
\textit{Human-curated skill}    & \shortstack[c]{$43.6$\\{\tiny $(+13.7)$}} & \shortstack[c]{$38.8$\\{\tiny $(+3.8)$}} & \shortstack[c]{$50.0$\\{\tiny $(+25.7)$}} & \shortstack[c]{$43.3$\\{\tiny $(+5.0)$}} & \shortstack[c]{$40.0$\\{\tiny $(+10.9)$}} & \shortstack[c]{$30.0$\\{\tiny $(+15.0)$}} & \shortstack[c]{$52.5$\\{\tiny $(+25.0)$}} & \shortstack[c]{$32.0$\\{\tiny $(+0.0)$}} & \shortstack[c]{$62.0$\\{\tiny $(+22.0)$}} & --- & --- & --- \\
\textit{Self-Gen} & \shortstack[c]{$32.0$\\{\tiny $(+2.1)$}}  & ---  & ---  & ---  & ---  & ---  & ---  & ---  & --- & --- & --- & --- \\
\sccontrol{}                             & \shortstack[c]{$33.9$\\{\tiny $(+4.0)$}}  & \shortstack[c]{$8.6$\\{\tiny $(-26.4)$}} & \shortstack[c]{$50.0$\\{\tiny $(+25.7)$}} & \shortstack[c]{$46.4$\\{\tiny $(+8.1)$}} & \shortstack[c]{$25.0$\\{\tiny $(-4.1)$}} & \shortstack[c]{$20.0$\\{\tiny $(+5.0)$}} & \shortstack[c]{$26.7$\\{\tiny $(-0.8)$}} & \shortstack[c]{$63.8$\\{\tiny $(+31.8)$}} & \shortstack[c]{$30.0$\\{\tiny $(-10.0)$}} & --- & --- & --- \\
\textbf{Evolver $R{=}1$}         & \shortstack[c]{$48.2$\\{\tiny $(+18.3)$}} & \shortstack[c]{$18.0$\\{\tiny $(-17.0)$}} & \shortstack[c]{$58.0$\\{\tiny $(+33.7)$}} & \shortstack[c]{$75.0$\\{\tiny $(+36.7)$}} & \shortstack[c]{$40.0$\\{\tiny $(+10.9)$}} & \shortstack[c]{$23.3$\\{\tiny $(+8.3)$}} & \shortstack[c]{$20.0$\\{\tiny $(-7.5)$}} & \shortstack[c]{$45.0$\\{\tiny $(+13.0)$}} & \shortstack[c]{$46.7$\\{\tiny $(+6.7)$}} & \shortstack[c]{$0.991$\\{\tiny $(-0.036)$}} & \shortstack[c]{$\mathbf{1.437}$\\{\tiny $(+0.320)$}} & \shortstack[c]{$2.185$\\{\tiny $(+0.859)$}} \\
\textbf{Evolver $R{=}2$}         & \shortstack[c]{$\mathbf{56.9}$\\{\tiny $(+27.0)$}} & \shortstack[c]{$\mathbf{50.4}$\\{\tiny $(+15.4)$}} & \shortstack[c]{$\mathbf{63.8}$\\{\tiny $(+39.5)$}} & \shortstack[c]{$\mathbf{83.9}$\\{\tiny $(+45.6)$}} & \shortstack[c]{$\mathbf{72.0}$\\{\tiny $(+42.9)$}} & \shortstack[c]{$\mathbf{40.0}$\\{\tiny $(+25.0)$}} & \shortstack[c]{$\mathbf{60.0}$\\{\tiny $(+32.5)$}} & \shortstack[c]{$\mathbf{84.0}$\\{\tiny $(+52.0)$}} & \shortstack[c]{$\mathbf{64.0}$\\{\tiny $(+24.0)$}} & \shortstack[c]{$\mathbf{1.089}$\\{\tiny $(+0.062)$}} & \shortstack[c]{$1.218$\\{\tiny $(+0.101)$}} & \shortstack[c]{$\mathbf{2.226}$\\{\tiny $(+0.900)$}} \\
\bottomrule
\end{tabular}
\vspace{-2pt}}
\end{table*}

A per-task decomposition against the human-curated skill at $R{=}2$ shows $24/83$ wins ($28.9\%$), $38/83$ ties ($45.8\%$), and $21/83$ losses ($25.3\%$); the cumulative fraction with \model\ $\geq$ human-curated is $62/83 = 74.7\%$. The headline gain comes primarily from \emph{widening the solvable set} rather than from dominating every task --- human-curated skills still win on roughly a quarter of tasks, typically those with a highly domain-specific DSL or convention where hand-written prose beats skills produced by the \model\ meta-skill (Appendix~\ref{app:cases}). A single self-generated skill (\sccontrol{}) does not clear the human-curated baseline; skills produced by the \model\ meta-skill do, and a single refinement iteration extends the lead to a double-digit margin.

\subsection{Component Ablation}
\label{sec:exp-ablation}

We ablate the second iteration of the loop ($R{=}1$ vs.\ $R{=}2$), holding model, task set, harness, anti-cheating layers, oracle policy and training variant constant. At $R{=}1$ the pipeline runs a single understand--explore--analyze--synthesize--audit pass with no refinement; at $R{=}2$ a second iteration redeploys $v_1$ as a real dependency on the training task, re-explores with $v_1$ live, and patches.

On the full $83$-task scope, the second iteration lifts aggregate avg@$5$ from $48.2\%$ to $56.87\%$ ($+8.7$\,percentage points), turning a $+4.6$\,percentage points edge over the curated baseline at $R{=}1$ into a $+13.3$\,percentage points edge at $R{=}2$. Refinement is responsible for roughly two thirds of the total gain over curated, not a marginal polish step.

Two case studies in Appendix~\ref{app:cases} clarify the mechanism: under-abstracted skills are repaired by hoisting the primary action into the skill header, and skills with the right prose but missing helper scripts are repaired by preserving discovery scripts during distillation. Both fixes generalised into authoring rules now enforced by Auditor Checks $8$ and $9$. Iterative re-exploration of the live $v_1$ exposes failure modes that one-pass distillation cannot see.

\subsection{Cost-Quality Trade-off}
\label{sec:exp-efficiency}

Table~\ref{tab:efficiency} summarizes both halves. \model\ at $R{=}2$ costs $\$3.92$ per task, only $\$0.28$ ($+8\%$) above $R{=}1$ for an $8.7$\,pp aggregate gain, and roughly half the $\$6.97$ per-task spend of \sccontrol{} --- a favourable point on the cost-quality frontier, well within the budget of an automated pipeline at scale ($\sim\$300$ for the $83$-task sweep). The evolved skill also accelerates the downstream agent: per validation trial we observe $-19.4\%$ tokens, $-15.3\%$ turns, and $-23.8\%$ wall-clock against a no-skill baseline on different task instances, indicating that the skill transfers methodology, not instance lookups. The same comparison for \sccontrol{} regresses on all three metrics, suggesting it adds prose without compressing downstream work. Refinement itself is also cheaper than first-pass exploration ($r{=}1$ vs $r{=}0$: $-6.0\%/-8.9\%/-6.9\%$). Skill compression and accuracy gains arrive together at $\sim\$4$ per task with a one-time $+8\%$ refinement overhead.

\begin{table}[t]
\centering
\small
\setlength{\tabcolsep}{5pt}
\caption{Per-trial cost and efficiency on the $83$-task scope. Authoring rows give total per-task pipeline cost. Training-side rows compare $r{=}0$ and $r{=}1$ exploration on $\mathcal{T}_{\text{train}}$. Validation-side rows compare no-skill exploration on $\mathcal{T}_{\text{train}}$ against with-skill validation on $\mathcal{T}_{\text{val}}$.}
\label{tab:efficiency}
\begin{tabular}{@{}l r r r@{}}
\toprule
\textbf{Phase} & \textbf{Tokens} & \textbf{Turns} & \textbf{Duration} \\
\midrule
\multicolumn{4}{@{}l}{\emph{Authoring (per-task pipeline cost, USD)}} \\
\quad \sccontrol{}                      & \multicolumn{3}{c}{$\$6.97$} \\
\quad \model\ ($R{=}1$)    & \multicolumn{3}{c}{$\$3.64$} \\
\quad \model\ ($R{=}2$)    & \multicolumn{3}{c}{$\$3.92$ ($+8\%$)} \\
\midrule
\multicolumn{4}{@{}l}{\emph{Training-side (within pipeline)}} \\
\quad $r{=}0$ (no skill)        & $299.3$k    & $10.1$    & $213$s    \\
\quad $r{=}1$ (with $v_1$)      & $281.4$k    & $9.2$     & $198$s    \\
\quad $\Delta$                  & $-6.0\%$    & $-8.9\%$  & $-6.9\%$  \\
\midrule
\multicolumn{4}{@{}l}{\emph{Validation-side (domain-skill agent)}} \\
\quad no-skill                  & $423.9$k    & $12.5$    & $201$s    \\
\quad with \sccontrol{} skill           & $525.4$k    & $14.6$    & $208$s    \\
\quad with evolved skill        & $341.9$k    & $10.6$    & $153$s    \\
\quad $\Delta$ (evolved vs no-skill) & $-19.4\%$   & $-15.3\%$ & $-23.8\%$ \\
\bottomrule
\end{tabular}
\end{table}

\subsection{Per-Category Analysis}
\label{sec:exp-analysis}

\begin{figure*}[!htbp]
\centering
\begin{minipage}[b]{0.58\textwidth}
\centering
\begin{tikzpicture}
\begin{axis}[
  ybar,
  width=\linewidth,
  height=4.0cm,
  bar width=4.5pt,
  ymin=0, ymax=110,
  ylabel={avg@$5$ (\%)},
  ylabel style={font=\scriptsize},
  symbolic x coords={A,B1,B2,B3,C1,C2,D},
  xtick=data,
  xticklabel style={font=\scriptsize},
  yticklabel style={font=\scriptsize},
  enlarge x limits=0.10,
  legend style={font=\scriptsize, at={(0.5,1.04)}, anchor=south, legend columns=4, draw=none, /tikz/every even column/.append style={column sep=4pt}},
  axis line style={-},
  tick style={black, thin},
  grid=major,
  major grid style={gray!18, dashed, thin},
  axis on top,
]
\addplot+[draw=black!70, fill=gray!50]   coordinates {(A,94)(B1,24)(B2,40)(B3,40)(C1,0)(C2,0)(D,0)};
\addplot+[draw=black!70, fill=green!45!black!50] coordinates {(A,89)(B1,64)(B2,40)(B3,14)(C1,65)(C2,23)(D,0)};
\addplot+[draw=black!70, fill=blue!25]   coordinates {(A,89)(B1,53)(B2,100)(B3,57)(C1,70)(C2,50)(D,20)};
\addplot+[draw=black!70, fill=blue!70]   coordinates {(A,89)(B1,65)(B2,100)(B3,47)(C1,78)(C2,30)(D,40)};
\legend{$A$ (no skill), $B$ (curated), Evolver ($R{=}1$), Evolver ($R{=}2$)}
\end{axis}
\end{tikzpicture}\\[2pt]
{\footnotesize (a) Bar chart}
\end{minipage}\hfill
\begin{minipage}[b]{0.40\textwidth}
\centering
\scriptsize
\setlength{\tabcolsep}{2.8pt}
\begin{tabular}{@{}l c c c c c c@{}}
\toprule
\textbf{Cat.} & \textbf{$n$} & \textbf{$A$} & \textbf{$B$} & \textbf{SC-SB} & \textbf{Ev.$_{1}$} & \textbf{Ev.$_{2}$} \\
\midrule
A   & $20$ & $.94$ & $.89$ & $.45$ & $.89$ & $.89$ \\
B1  & $14$ & $.24$ & $.64$ & $.38$ & $.53$ & $\mathbf{.65}$ \\
B2  & $2$  & $.40$ & $.40$ & $\mathbf{1.00}$ & $\mathbf{1.00}$ & $\mathbf{1.00}$ \\
B3  & $7$  & $.40$ & $.14$ & $.38$ & $\mathbf{.57}$ & $.47$ \\
C1  & $11$ & $.00$ & $.65$ & $.11$ & $.70$ & $\mathbf{.78}$ \\
C2  & $6$  & $.00$ & $.23$ & $.07$ & $\mathbf{.50}$ & $.30$ \\
D   & $23$ & $.00$ & $.00$ & $.25$ & $.20$ & $\mathbf{.40}$ \\
\bottomrule
\end{tabular}\\[2pt]
{\footnotesize (b) Numerical values}
\end{minipage}
\caption{Per-category avg@$5$ across the SkillsBench skill-utility taxonomy. Evolver wins biggest where curated skills hurt ($B3$) or fail entirely ($C$ and $D$ categories). On the $A$ bucket the agent already solves the task without a skill, so the pipeline is not invoked and the bar repeats the no-skill rate. Categories: $A$ = already easy ($n{=}20$), $B1{/}B2{/}B3$ = curated helps$/$is neutral$/$hurts, $C1{/}C2$ = curated unlocks (strong$/$weak), $D$ = neither baseline nor curated solves it. \sccontrol{} (abbreviated SC-SB in column headers) shown for tasks where the Anthropic skill-creator adaptation was run.}
\label{fig:per-category}
\end{figure*}
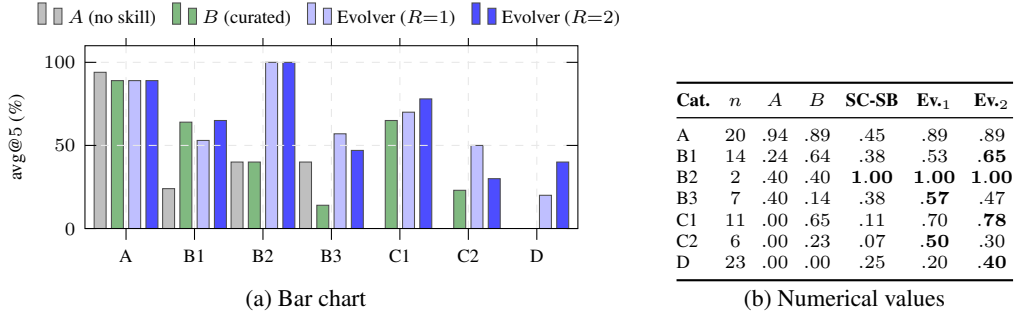

To understand where the gains come from, we project the per-task results onto the SkillsBench skill-utility taxonomy: $A$ = the agent already solves the task without help; $B1{/}B2{/}B3$ = curated skill helps$/$is neutral$/$hurts; $C1{/}C2$ = curated skill unlocks the task strongly$/$weakly; $D$ = neither baseline nor curated skill solves it (Figure~\ref{fig:per-category}).

The picture is highly non-uniform. Largest gains: $B2$ ($+60$\,pp over $B$, only two tasks --- high variance), $D$ ($+40$\,pp from a $0\%$ floor), $B3$ ($+33$\,pp over a curated baseline that actively hurts), $C1$ ($+13$\,pp). On $B1$, where curated already helps, Evolver adds a small marginal lift; on $A$ it matches no-skill. Refinement adds the most on $D$ ($+20$\,pp over $R{=}1$) and $C1$ ($+8$\,pp). Gains concentrate where curated fails hardest --- the right asymmetry for a meta-skill targeting that gap.

\textbf{Failure modes.} Remaining failures cluster into three classes. \emph{(H1) Pipeline bugs} --- under-abstraction, discovery-script loss, silent-bypass, train/val schema drift --- now addressed by Auditor Checks $7$--$9$. \emph{(H2) Train/validation domain gap} --- training variants drift in bug family, schema, or parameter range; we treat this as realistic in-domain shift. \emph{(H3) Model-capacity walls} --- e.g., whisper-large-v3 on CPU FP$32$ cannot meet the speaker-diarization gate (the curated oracle itself fails on our hardware).

\section{Conclusion and Limitations}
\label{sec:conclusion}
\label{sec:limitations}

\model shows that an agent can acquire a reusable procedural skill from a bounded set of deployment-time trials without updating model weights. The key is to close the loop around deployed skill use: strategy-diversified exploration sends the current skill to fresh Domain-Skill Agents, contrastive updates turn high- versus low-reward traces into localized artifact edits, and an independent Auditor gates whether the revised skill can enter the accepted sequence. This reframes skill creation from one-shot authoring into artifact-level adaptation for Domain-Skill Agent.

\textbf{Single-LLM evaluation.} The meta-skill is loaded through the same CLI-agent interface as any domain skill, so it is in principle LLM- and agent-agnostic --- our spot tests on GPT + Codex run end-to-end. A full benchmark sweep on alternative SOTA LLMs, however, was not run due to cost, and the headline numbers therefore reflect a single configuration, Claude Opus 4.6 + Claude Code; cross-LLM parity is a design property rather than a benchmark-scale empirical finding.

\textbf{Refinement depth not characterized.} The cap $R{=}2$ is a compute-budget choice rather than a measured optimum: our $R{=}1{\to}R{=}2$ ablation (\S\ref{sec:exp-ablation}) shows the second iteration is responsible for roughly two-thirds of the gain over the curated baseline, suggesting non-trivial margin may remain at $R{\geq}3$. We did not run deeper sweeps at scale.

\textbf{Benchmark coverage.} \model is evaluated on SkillsBench ($83$ tasks, binary) and a small KernelBench probe ($3$ tasks, scalar speedup), reflecting the youth of skill-authoring evaluation.

\textbf{Richer process-level signals.} The \model\ Agent already consumes basic execution-efficiency information (per-trial tokens, turns, wall-clock; Table~\ref{tab:efficiency}) when contrasting traces; what remains open is whether richer process signals --- step-level grounding, intermediate verifier checks, latency-on-critical-path --- can be folded into the contrast to further lift skill quality, an axis we do not characterize here.

\textbf{Single-task scope; no skill library.} \model targets one newly-arrived task at a time and produces an artifact for that task; we do not address how a population of such artifacts is organized or maintained. Cross-task reuse, library deduplication, and parent--sibling specialization across related tasks --- the maintenance side of the skill lifecycle --- remain open for future work.

\newpage
\bibliographystyle{iclr2026_conference}
\bibliography{references}

\appendix
\section{Appendix}
\label{sec:appendix}

% TODO: prose by section-author
%
% Appendix hooks (from paper-outline.md):
%  - A. SkillsBench task list and categorization.
%  - B. Training-variant generation methodology.
%  - C. SkillEvolver (N=1) / SkillEvolver (N=2) skill excerpts.
%  - D. Per-task result tables.
%  - E. Failure-mode case studies.

\subsection{\model\ Full Pseudocode}
\label{app:evolver-alg}

Figure~\ref{fig:pipeline} in the main text gives the high-level pipeline; Algorithm~\ref{alg:evolver-a} below makes the strategy-diversified sampling and deploy-then-refine loop explicit.

\begin{algorithm}[!t]
\small
\SetKwFunction{Parse}{Parse}
\SetKwFunction{Diverse}{DiverseStrategies}
\SetKwFunction{Explore}{Explore}
\SetKwFunction{Contrast}{Contrast}
\SetKwFunction{Distill}{Distill}
\SetKwFunction{Patch}{SurgicalPatch}
\SetKwFunction{Auditor}{Auditor}
\SetKwFunction{Validate}{Validate}
\SetKwFunction{Score}{score}
\SetKwInOut{Input}{Input}
\SetKwInOut{Output}{Output}
\DontPrintSemicolon

\Input{task $\mathcal{T}$; iteration cap $R$; explore width $K$; validate trials $V$}
\Output{$\pi(v^{*};\mathcal{T}_{\text{val}})$}

axes $\leftarrow$ \Parse{$\mathcal{T}_{\text{train}}$} \tcp*{\S\ref{sec:method-metaskill}\ \ one-shot Understand}
$\mathcal{S}_0 \leftarrow$ \Diverse{axes,$\varnothing$,$\varnothing$} \tcp*{\S\ref{sec:method-sampling}\ bootstrap: $K$ strong-prior strategies routed by trial index}
$\tau_0 \leftarrow$ \Explore{$\mathcal{T}_{\text{train}},\mathcal{S}_0,v{=}\varnothing,K$} \tcp*{\S\ref{sec:method-sampling}\ $K$ parallel trials; trial $i$ loads $s_{0,i}$ via env index}
$\Delta_0 \leftarrow$ \Contrast{$\tau_0^{+},\tau_0^{-}$} \tcp*{\S\ref{sec:method-synthesize}\ winners $\setminus$ losers}
$v_1 \leftarrow$ \Distill{$\Delta_0$}; \textbf{mirror} $v_1$ to \texttt{output/} \tcp*{\S\ref{sec:method-synthesize}\ failsafe deploy copy}
\For(\tcp*[f]{\S\ref{sec:method-sampling}\ deploy-then-stress-test loop}){$r \leftarrow 1$ \KwTo $R{-}1$}{
  \textbf{deploy} $v_r$ as a live skill in the trial container \tcp*{\S\ref{sec:method-sampling}\ \textbf{real-scenario sampling}}
  $\mathcal{S}_r \leftarrow$ \Diverse{axes,$v_r,\tau_{r{-}1}$} \tcp*{\S\ref{sec:method-sampling}\ \textbf{persists:} $K$ stress-test priors aimed at $v_r$'s weak spots}
  $\tau_r \leftarrow$ \Explore{$\mathcal{T}_{\text{train}},\mathcal{S}_r,v_r,K$} \tcp*{\S\ref{sec:method-sampling}\ trial $i$ loads $s_{r,i}$ first, then consults $v_r$}
  $\Delta_r \leftarrow$ \Contrast{$\tau_r^{+},\tau_r^{-}$} \tcp*{\S\ref{sec:method-synthesize}\ where did $v_r$ mislead?}
  $v_{r{+}1} \leftarrow$ \Patch{$v_r,\Delta_r$} \tcp*{\S\ref{sec:method-synthesize}\ surgical, not rewrite}
  \lIf(\tcp*[f]{\S\ref{sec:method-audit}\ continue-or-exit}){\Auditor{$v_{r{+}1},\mathcal{T}_{\text{train}},\tau_r$} clean $\land$ \#pass$(\tau_r)\geq 3K/4$}{\textbf{break}}
}
$v^{*} \leftarrow \arg\!\max_{v\in\{v_1,\ldots,v_R\}} \Score{$v;\mathcal{T}_{\text{train}}$}$ \tcp*{\S\ref{sec:method-metaskill}\ Finalize (no Harbor call)}
\Return \Validate{$v^{*},\mathcal{T}_{\text{val}},V$} \tcp*{\S\ref{sec:method-metaskill}\ held-out validation}
\caption{\textbf{\model}. The key design choice is
\emph{strategy-diversified sampling on a deployed skill}: every
iteration writes $K$ strong-prior strategies (line~2 at $r{=}0$,
line~8 at $r{>}0$) that Harbor routes to parallel trials via a
per-trial environment index (lines~3 and~9); from $r{=}1$ onwards
those strategies are aimed at the current candidate skill $v_r$'s
observed weak spots rather than at the original decision axes.
Line~7 is the second half of the commitment: $v_r$ is installed as
a real dependency in the trial container, so the contrast at
line~10 reflects where a fresh using-agent was actually helped or
misled by the deployed skill. Line~11 applies a surgical patch
rather than rebuilding from scratch, and line~12 invokes the
fresh-session auditor from \S\ref{sec:method-audit}.}
\label{alg:evolver-a}
\end{algorithm}

\subsection{Auditor check list}
\label{app:auditor}

The Auditor subagent (\S\ref{sec:method-audit}) runs the nine mechanical checks in Table~\ref{tab:auditor}. Checks~1--6 cover standard content-level leakage patterns; Checks~7--9 are specific to the deployed-skill regime this paper introduces (parametric-axis under-abstraction, primary-action hoisting, silent-bypass) and are detectable only because the refinement signal comes from a deployed skill's handoff traces rather than from the authoring agent's reflection on its own work.

\begin{table}[t]
\centering
\scriptsize
\setlength{\tabcolsep}{3pt}
\caption{Auditor checks. $^{\star}$ marks critical checks; any hit forces a targeted patch in the next iteration. Checks 1--6 target content-level overfit (standard leakage patterns). Checks 7--9 target deployed-skill regime failures we introduce: parametric-axis under-abstraction, structural failure to hoist the primary action, and silent-bypass at runtime. Checks 7--9 are observable only because our pipeline refines against traces of the candidate skill as a live dependency (\S\ref{sec:method-sampling}), not against the authoring agent's reflection on its own work.}
\label{tab:auditor}
\begin{tabular}{@{}llp{4.3cm}@{}}
\toprule
\# & Check & What it catches \\
\midrule
1  & Framing$^{\star}$       & \texttt{name}/\texttt{description} borrows a training-instance business noun instead of the abstract operation. \\
2  & Literals$^{\star}$      & Hardcoded training filenames, field names, entity strings, or soft-qualifier numericals (``\texttt{typically} $<$ 2.5''). \\
2b & Script bloat            & Individual script exceeds 200 lines (important) / 400 lines (critical) -- almost always a bundled solution generator, not a focused utility. \\
3  & Untraceable             & Imperative assertions (``use X not Y'', ``never'', ``required'') without trace provenance and not pretraining-obvious. \\
4  & Shape-bake$^{\star}$    & Scripts index by hardcoded column, sheet, or key without a runtime probe (\texttt{df.columns}, \texttt{wb.sheetnames}); $\geq 3$ \texttt{if "keyword" in desc} branches in one script flag keyword-dispatch solution-generators. \\
5  & Coverage                & Mechanical task with zero bundled scripts (skipped in high-pass mode). \\
6  & X-ref$^{\star}$         & Any $\geq 4$-char string literal in skill files matching a training filename, field, or value. \\
7  & Under-abstraction$^{\star}$ & Imperative claims embedding a parametric-axis constant (e.g.\ terminal-state code, threshold) without a sibling ``re-derive at runtime'' instruction or invariance provenance. \\
8  & Primary-action hoisting$^{\star}$ & Skill declares a \texttt{primary\_script} but \texttt{SKILL.md} routes constraints/background prose \emph{before} the invocation block, so the using-agent reads constraints first and never invokes the script. \\
9  & Silent-bypass$^{\star}$ & Skill declares a \texttt{primary\_script} but exploration traces contain zero Bash invocations of it under majority-fail -- a correct skill is quietly ignored at runtime. \\
\bottomrule
\end{tabular}
\end{table}

\subsection{Contamination controls}
\label{app:anticheat}
Applied identically across \model\ (at both $R{=}1$ and $R{=}2$) and \sccontrol{}.

\paragraph{Layer 1: train/test split.} All iterations of the evolve loop run on $\mathcal{T}_{\text{train}}$, a generated variant with different data, filenames, and sometimes sub-domain from $\mathcal{T}_{\text{val}}$; validation runs on $\mathcal{T}_{\text{val}}$. A skill that encodes a training-specific filename or value silently fails validation -- the file is not there. Before each exploration run, the curated training skill is deleted at source, so it is never reachable. Contamination discipline in this form follows precedent in LLM evaluation~\citep{sainz2023contamination}; the agent setting adds a new surface, since the authoring agent may itself encode training-label content into the skill artifact.

\paragraph{Layer 2: workspace whitelist.} A PreToolUse hook denies every agent tool call outside a single per-run workspace prefix. The validation task directory, curated validation skill, and test suites live outside and are unreachable. A denylist tripwire checked before the whitelist denies \texttt{..} traversal and any path resolving into the curated training-skill slot. Path resolution checks both the raw and symlink-resolved paths.

\subsection{\sccontrol{} (Control Baseline) Algorithm and Alignment Table}
\label{app:scsb}

Algorithm~\ref{alg:sc-skillsbench} gives the pseudocode for \sccontrol{}, the canonical Anthropic skill-creator~\citep{anthropic2025skillcreator, anthropic2025agentskills} adapted to remove the human: Eval Designer, Grader, and Analyzer subagents replace the Capture Intent interview, human grading, and eval-viewer feedback. Table~\ref{tab:alignment} aligns \model\ (at both $R{=}1$ and $R{=}2$) and \sccontrol{} on trial counts, budgets, oracle policy, isolation, and authoring mechanism.

\begin{algorithm}[!t]
\scriptsize
\SetKwFunction{Spawn}{spawn\_subagent}
\SetKwFunction{Draft}{draft}
\SetKwFunction{ABTest}{run\_ab\_eval}
\SetKwFunction{Improve}{improve}
\SetKwFunction{Validate}{harbor\_validate}
\SetKwInOut{Input}{Input}\SetKwInOut{Output}{Output}
\DontPrintSemicolon

\Input{$\mathcal{T}=(\mathcal{T}_{\text{train}},\mathcal{T}_{\text{val}})$, iteration cap $J{=}2$, validation trials $V{=}5$}
\Output{Pass@$V$ on $\mathcal{T}_{\text{val}}$}

$E \leftarrow$ \Spawn{eval\_designer; reads $\mathcal{T}_{\text{train}}$ instr.\ + oracle + skill-creator guide} \tcp*{Phase 1: replaces human Capture Intent}
$v_1 \leftarrow$ \Draft{$\mathcal{T}_{\text{train}}, E$, skill-creator guide} \tcp*{Phase 2: main session}
\For{$r \leftarrow 1$ \KwTo $J$}{
  $O_r \leftarrow$ \ABTest{$v_r,\mathcal{T}_{\text{train}}$; \{with,\,without\}} \tcp*{Phase 3: \emph{local} subprocess A/B (no Harbor)}
  \ForEach(\tcp*[f]{Phase 4: replaces human grading; cannot read skill source}){$c\in\{\text{with},\text{without}\}$}{
    $g_{r,c}\leftarrow$ \Spawn{grader($O_{r,c},E$)}\;
  }
  $f_r \leftarrow$ \Spawn{analyzer($O_r,g_r$)} \tcp*{Phase 6: replaces eval-viewer feedback; cannot read skill source}
  $v_{r+1} \leftarrow$ \Improve{$v_r,f_r$} \tcp*{Phase 7: main session, follows upstream guide}
}
\Return \Validate{$v_{J+1},\mathcal{T}_{\text{val}},V$} \tcp*{Phase 9: only Harbor invocation in this pipeline}
\caption{\textbf{\sccontrol{} (control baseline).} Anthropic's skill-creator with three human seats replaced by isolated subagents (Eval Designer, Grader, Analyzer); the Improver remains in the main session. A/B self-tests run as local subprocess SDK sessions on $\mathcal{T}_{\text{train}}$; only Phase 9 validation invokes Harbor. By construction the Eval Designer reads $\mathcal{T}_{\text{train}}$ in full (including the test/solve oracle); Grader and Analyzer cannot read the skill source or $\mathcal{T}_{\text{train}}$.}
\label{alg:sc-skillsbench}
\end{algorithm}

% ------------------------------------------------------------------
% Alignment table
% ------------------------------------------------------------------
\begin{table*}[t]
\centering
\footnotesize
\setlength{\tabcolsep}{4pt}
\caption{Alignment of \model\ at $R{=}2$, \model\ at $R{=}1$ (the non-refining ablation), and \sccontrol{} across key design axes. All three pipelines share the same two-layer anti-cheating design (train/test split, workspace whitelist).}
\label{tab:alignment}
\begin{tabular}{@{}p{0.18\linewidth} p{0.22\linewidth} p{0.20\linewidth} p{0.28\linewidth}@{}}
\toprule
\textbf{Property} & \textbf{\model\ ($R{=}2$)} & \textbf{\model\ ($R{=}1$ ablation)} & \textbf{\sccontrol} \\
\midrule
Explore trials $K$ & 4 (\texttt{EVOLVER\_N\_\allowbreak EXPLORATION}) & 4 (\texttt{EVOLVER\_N\_\allowbreak EXPLORATION}) & 0 Harbor (local subprocess only) \\
Re-explore trials at $r{=}1$ & 4 (\texttt{EVOLVER\_N\_REFINE}) & --- (no refinement) & 0 Harbor (local A/B via subprocess) \\
Validate trials $V$ & 5 (\texttt{EVOLVER\_N\_\allowbreak VALIDATION}) & 5 (\texttt{EVOLVER\_N\_\allowbreak VALIDATION}) & 5 (hard-coded in \texttt{run\_and\_wait.py}) \\
Harbor invocations & 3 (explore + re-explore + validate) & 2 (explore + validate) & 1 (validate only) \\
Total Harbor trials & 13 ($2K{+}V$) & 9 ($K{+}V$) & 5 \\
Dollar budget & \$15 (\texttt{EVOLVER\_\allowbreak MAX\_BUDGET}) & \$15 (\texttt{EVOLVER\_\allowbreak MAX\_BUDGET}) & \$15 (\texttt{SCSB\_\allowbreak MAX\_BUDGET}) \\
Turn budget & 200 (\texttt{EVOLVER\_\allowbreak MAX\_TURNS}) & 200 (\texttt{EVOLVER\_\allowbreak MAX\_TURNS}) & 200 (\texttt{SCSB\_\allowbreak MAX\_TURNS}) \\
Oracle policy & Uncertainty-based: read \texttt{test\_outputs.py} only when traces leave gap; escalate to \texttt{solve.sh} only if still unclear & Same as evolver & Eval Designer reads \texttt{train-context/} in full (allowed by construction); Grader/Analyzer cannot read \texttt{train-context/} \\
Isolation mechanism & Path guard whitelist + train/test split + denylist tripwire & Same & Same ($+$ subagent narrow whitelists) \\
Iteration type & $r{=}2..R$ ($R{=}2$) refinement loop & Single iteration (no loop) & $r{=}1..J$ ($J{=}2$) authoring loop \\
Termination condition & Hard cap $R{-}1$ refinement iterations & n/a & Hard cap $J$ iterations \\
Finalize phase? & Yes (Phase 8: agent reasoning picks $v_1/v_R$/merge) & No (Phase 4 output = deployed skill) & No (Phase 8 = promote iteration-$J{+}1$ directly) \\
Deploys prior skill as real skill? & Yes (Phase 5: $v_{r-1}$ into \texttt{environment/}\allowbreak\texttt{skills/}) & No & No (local sessions, not Harbor) \\
A/B self-test mechanism & None (binary reward via Harbor) & None & \texttt{run\_ab\_eval.py} subprocess SDK sessions, no Harbor \\
Source for skill guide & \texttt{skill-evolver/}\allowbreak\texttt{SKILL.md} & \texttt{skill-evolver/}\allowbreak\texttt{SKILL.md} ($R{=}1$ config) & \texttt{skill-creator-sb/}\allowbreak\texttt{SKILL.md} + \texttt{skill-creator-upstream/}\allowbreak\texttt{SKILL.md} \\
\bottomrule
\end{tabular}
\end{table*}

\subsection{SkillsBench Task List and Categorization}
\label{app:tasks}

The 83 runnable SkillsBench tasks span 15$+$ domains: web development, data science, DevOps, chemistry, quantum computing, seismology, finance, document processing, ML infrastructure, control systems, game analytics, audio/video processing, security, scheduling, and more. Full per-task domain assignments, categorical labels ($A$, $B1$, $B2$, $B3$, $C1$, $C2$, $D$), and runnability status are released with the benchmark. Four tasks are excluded from our sweep: two require paid external APIs and two exhibit persistent infrastructure instability under our Harbor configuration; exclusion is symmetric across conditions, leaving the $83$-task paper scope.

\subsection{Training-Variant Generation Methodology}
\label{app:variants}

Training variants $\mathcal{T}_{\text{train}}$ are disjoint from the corresponding $\mathcal{T}_{\text{val}}$ on filenames and data values, sometimes on sub-domain. Variants are generated per task by a separate pipeline that preserves task structure (the same test specification format, the same expected output shape) while resampling inputs; the curated training skill is deleted at source, so it is never reachable during exploration. The generation pipeline is orthogonal to \model\ (at any $R$); we use its output as-is.

\subsection{Per-Task Result Tables and Finalize Choice Distribution}
\label{app:per-task}

\begin{figure*}[!htbp]
\centering
\includegraphics[width=\textwidth]{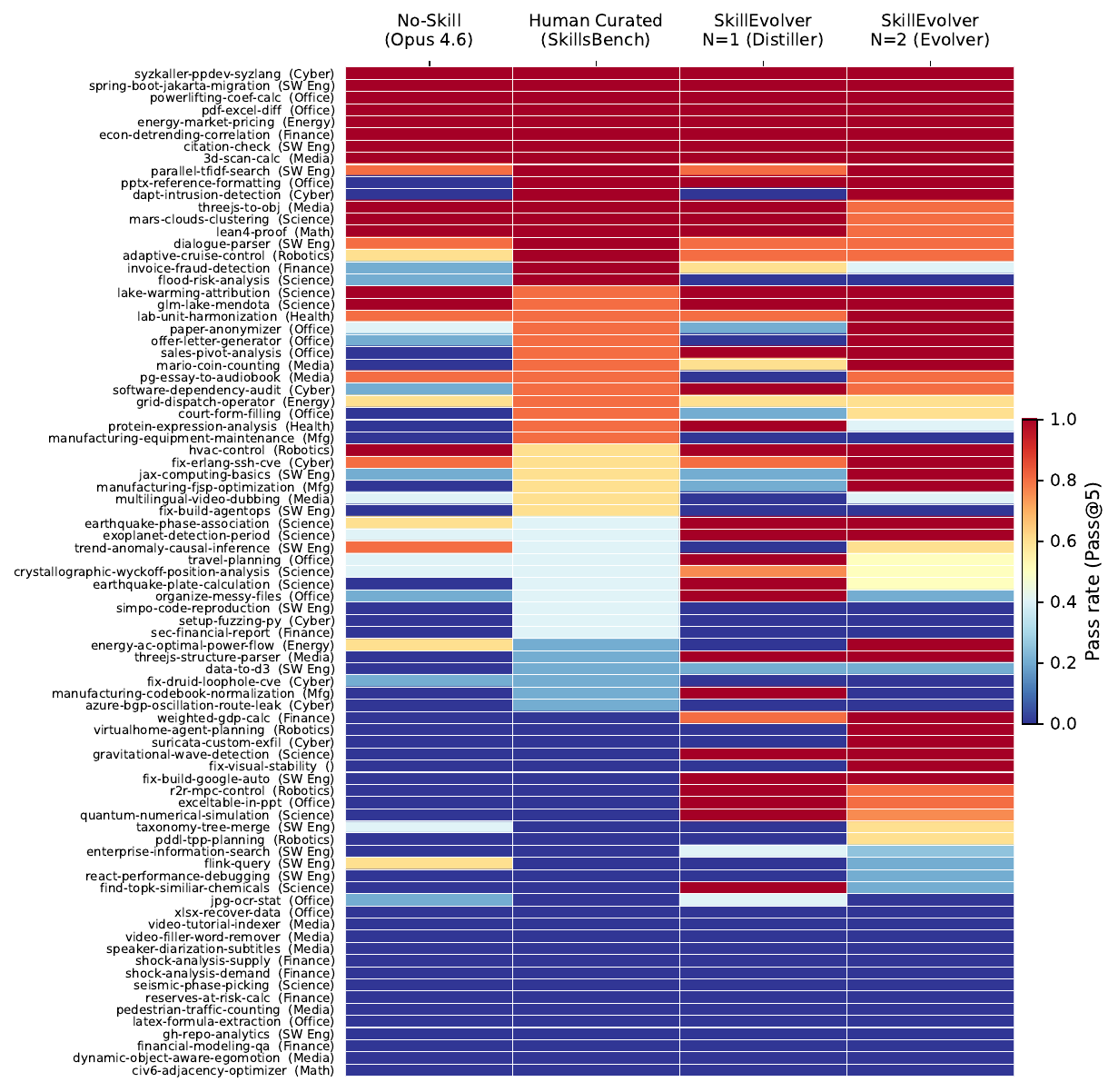}
\caption{Per-task Pass@$5$ on the 83-task paper scope under four Opus~4.6 conditions. Rows sorted by Curated descending. \textbf{No-Skill}: Opus~4.6 with no skill installed. \textbf{Human Curated}: the SkillsBench curated skill. \textbf{\model\ $R{=}1$}: the non-refining ablation. \textbf{\model\ $R{=}2$}: the full Evolver loop (\S\ref{sec:method-metaskill}).}
\label{fig:appendix-heatmap}
\end{figure*}

Per-task Pass@5 for $A$, $B$, \sccontrol{}, and \model\ at both $R{=}1$ and $R{=}2$ appears in the released results database. For \model, per-task Finalize choice ($v_1$ vs $v_2$ vs merge), per-phase Pass@5 (exploration, refinement, validation), and per-task token and dollar cost are recorded for every run. Figure~\ref{fig:appendix-heatmap} visualizes the per-task Pass@5 across the four conditions, in the same heatmap form as Figures~11--12 of the SkillsBench paper. Table~\ref{tab:categories} gives the per-category Pass@5 breakdown and Table~\ref{tab:authoring-cost} the skill-authoring cost comparison, both referenced from \S\ref{sec:exp-main}.

\begin{table}[!ht]
\centering
\footnotesize
\setlength{\tabcolsep}{3pt}
\caption{Skill-authoring cost and Harbor-trial footprint, including the $V{=}5$ validation trials. \sccontrol{} authoring runs in local subprocess sessions; only its $V{=}5$ validation trials hit Harbor.}
\label{tab:authoring-cost}
\begin{tabular}{@{}l c c@{}}
\toprule
\textbf{Condition} & \textbf{Cost (\$)} & \textbf{Harbor trials} \\
\midrule
\sccontrol{}                                  & \$6.97          & 5 ($V$ only)    \\
\model\ ($R{=}1$ ablation)       & \$3.64          & 9 ($4{+}5$)     \\
\textbf{\model\ ($R{=}2$, ours)} & \textbf{\$3.92} & 13 ($4{+}4{+}5$) \\
\bottomrule
\end{tabular}
\end{table}

\begin{table}[!ht]
\centering
\scriptsize
\setlength{\tabcolsep}{3pt}
\caption{Per-category Pass@$5$ breakdown by SkillsBench category (B1 curated helps, B2 curated neutral, B3 curated hurts, C1 skill-unlocked strong, C2 skill-unlocked weak, D hopeless). Values reproduce Figure~\ref{fig:per-category}(b) for ease of reference.}
\label{tab:categories}
\begin{tabular}{@{}l c c c c c c@{}}
\toprule
\textbf{Cat.} & \textbf{$n$} & \textbf{$A$} & \textbf{$B$} & \textbf{SC-SB} & \textbf{Ev.\,$R{=}1$} & \textbf{Ev.\,$R{=}2$} \\
\midrule
B1 & $14$ & $.24$ & $.64$ & $.38$            & $.53$           & $\mathbf{.65}$ \\
B2 & $2$  & $.40$ & $.40$ & $\mathbf{1.00}$  & $\mathbf{1.00}$ & $\mathbf{1.00}$ \\
B3 & $7$  & $.40$ & $.14$ & $.38$            & $\mathbf{.57}$  & $.47$ \\
C1 & $11$ & $.00$ & $.65$ & $.11$            & $.70$           & $\mathbf{.78}$ \\
C2 & $6$  & $.00$ & $.23$ & $.07$            & $\mathbf{.50}$  & $.30$ \\
D  & $23$ & $.00$ & $.00$ & $.25$            & $.20$           & $\mathbf{.40}$ \\
\bottomrule
\end{tabular}
\end{table}

\subsection{Case Study of SkillsBench}
\label{app:cases}

We discuss six representative cases --- three lifts and three failures --- to illustrate the mechanisms behind the aggregate numbers in \S\ref{sec:experiments}. Each case names the bug or mechanism we identified from the traces, not just the outcome.

\paragraph{Positive 1: \texttt{manufacturing-fjsp-optimization} ($0.2 \to 1.0$ at $R{=}2$).} The $v_1$ skill listed subtask recipes but did not hoist the one-shot primary action. Refinement traces showed trials stalling on ``which script do I invoke first?''. $v_2$ promoted the primary script to the top of the skill --- \emph{Primary-Action Hoisting} --- and trials completed end-to-end. The fix generalised into Auditor Check~$8$.

\paragraph{Positive 2: \texttt{paper-anonymizer} ($0.2 \to 1.0$ at $R{=}2$).} The $v_1$ prose prescribed the right approach (PyMuPDF-based redaction with handling for unicode quote variants and rotated arXiv sidebar text) but the underlying \texttt{inspect\_pdf.py} discovery helper was not bundled into \texttt{scripts/}, so $r{=}1$ trials could not execute the prescribed strategy. $v_2$ preserved the helper --- \emph{Discovery-Script Preservation} --- and trials completed end-to-end. The fix generalised into Auditor Check~$9$.

\paragraph{Positive 3: \texttt{virtualhome-agent-planning} ($0.0 \to 1.0$ at $R{=}2$).} The $v_1$ skill body was correct (generic \texttt{pyperplan}-CLI invocation that emits the parenthesized PDDL plan form required by \texttt{unified\_planning.io.PDDLReader}) but the \texttt{description:} frontmatter did not trigger Skill-tool invocation in Claude Code: trials hand-wrote plans in the functional-form syntax shown in the task instruction example, which \texttt{PDDLReader} rejects with \texttt{UPException}. $v_2$ rewrote the description to name the SkillsBench task and cite the \texttt{UPException} trap, raising the Skill-tool invocation rate to $5/5$. This is a description-level fix with no body-level changes.

\paragraph{Negative 1: \texttt{court-form-filling} (no-skill $4/4$, $R{=}1$ and $R{=}2$ both $0/5$).} The cleanest negative-transfer case in the sweep. Training is a medical intake form (\texttt{intake-blank.pdf}); validation is a California Small Claims SC-$100$. Library-level knowledge transferred correctly --- $43/47$ verifier sub-tests still pass with the skill (pypdf API, \texttt{/Yes} checkbox encoding, page-ordering via \texttt{writer.append(reader)}). The four failing sub-tests all match one pattern: SC-$100$ has \emph{default-no} checkboxes that must be \emph{actively} checked when the case description does not mention the topic. The distilled skill encoded the heuristic ``only fill fields mentioned in the description; leave unmentioned fields empty,'' which is true for medical intake and false for legal forms; refinement re-explored the same training data and baked the heuristic deeper. This motivates \emph{multi-domain training} as future work: distilling from a single-domain training variant gives refinement no signal to abstract away a domain-specific decision rule.

\paragraph{Negative 2: \texttt{invoice-fraud-detection} ($R{=}1$ $0.6 \to N{=}2$ $0.4$).} A regression-on-refinement case. Multiple independent failure modes in the validation rubric: the $v_1$ skill correctly captured the \texttt{po\_number}-must-be-\texttt{null}-when-missing convention and the five-tier fraud check priority order, but refinement re-exploration locked onto the same single failure mode that $v_1$ fixed and over-hardened it, while a partial-string-matching pitfall (\texttt{partial\_ratio} returns $100$ for ``Vendor 1'' vs ``Vendor 11'') remained underspecified in $v_2$. Iterative refinement is beneficial when failure modes are independent, but can over-fit when the agent treats the largest visible failure as the whole problem.

\paragraph{Negative 3: \texttt{pptx-reference-formatting} (explore $1/4$, val $0/3$).} A failure-focused-analysis pathology. Failing exploration traces all lacked \texttt{a:buAutoNum} (auto-numbered bullets), and the agent correctly identified this and wrote a detailed skill covering it. But the passing trace also set \texttt{pPr.set('algn', 'ctr')} for paragraph centre alignment --- a detail buried at line~$634$ of a $926$-line trace. The distilled skill covered shape positioning (EMU coordinates) but not text alignment, and validation failed all three trials on \texttt{test\_titles\_center\_aligned} (expected \texttt{ctr}, got \texttt{l}). The diff-oriented analysis missed a non-trivial operation present in the passing trace because it did not correspond to an explicit failure mode in the failing traces. This is the strongest argument for an explicit \emph{passing-trace coverage} check during distillation, not just a failure-cause check.

\subsection{Case Study of KernelBench}
\label{app:cases-kb}

KernelBench serves a different role from the main SkillsBench sweep. The tasks are not discrete workflow-completion problems but continuous optimization problems, and the relevant signal is correctness-weighted speedup rather than Pass@$5$. We therefore use these runs to examine \emph{what kind of optimization knowledge the evolved skill captures} when the reward is scalar and architecture-sensitive.

\paragraph{Cross-architecture transfer via procedural optimization heuristics.} The three tasks in the main-text extension --- \texttt{deepnarrowmlp}, \texttt{shufflenet}, and \texttt{gru} --- span MLP, CNN, and RNN settings respectively. Under the continuous-reward \model\ setup, the evolved skill improves mean reward on all three at $R{=}2$: \texttt{deepnarrowmlp} from $1.027$ to $1.089$, \texttt{shufflenet} from $1.117$ to $1.218$, and \texttt{gru} from $1.326$ to $2.226$. The common pattern is not memorization of a single kernel trick, but acquisition of reusable optimization procedure: classify the architecture, identify the dominant compute bottleneck, and then choose architecture-specific implementation tactics. In this sense the learned artifact behaves less like a task answer and more like an optimization playbook.

\paragraph{What the skill appears to learn.} Inspection of the final \texttt{kernel-optim} lineage suggests three reusable knowledge types. First, it preserves \emph{decision rules}: distinguish matmul-heavy MLP/CNN workloads from recurrent sequence models and avoid applying the same precision or compiler choice blindly across them. Second, it preserves \emph{execution heuristics}: prefer concrete implementation interventions such as layout changes, kernel fusion opportunities, cuDNN-backed fast paths, or graph capture only when the traces indicate they are stable. Third, it preserves \emph{negative lessons}: failed precision settings, unstable compilation choices, and architecture-specific anti-patterns are retained as explicit constraints, which is especially important in continuous optimization because many partially-correct ideas still receive non-zero reward and would otherwise be hard to filter out.

\paragraph{Why \texttt{gru} benefits the most.} The largest gain appears on \texttt{gru} ($1.326 \to 2.226$). The traces suggest that recurrent-model optimization benefits disproportionately from a reusable skill because the space of superficially plausible but low-yield interventions is large: precision changes, compiler switches, and sequence-handling choices can all look promising locally. Once the skill accumulates explicit RNN-specific guidance and failed-attempt constraints, the using-agent wastes less search on those dead ends and reaches the stronger implementation regime much more reliably. This is exactly the kind of benefit one would expect if the skill is compressing optimization methodology rather than memorizing a benchmark-specific answer.

\paragraph{Why the smaller gains still matter.} The gains on \texttt{deepnarrowmlp} and \texttt{shufflenet} are smaller in absolute terms, but they are still informative. These tasks start from stronger no-skill baselines, so the remaining headroom is narrower. A modest positive shift therefore suggests that the evolved skill is not merely over-specialized to the most favorable task in the chain; it still helps the using-agent choose slightly better optimization actions even when the baseline agent is already reasonably competent. This is the main reason we interpret KernelBench as evidence of cross-scenario generalization rather than as a single-task anecdote.

\paragraph{A caution on comparability.} The KernelBench results should not be over-read as a second main benchmark on par with SkillsBench. The sample is small because these tasks require real GPU kernel evaluation, the metric is different, and the bookkeeping of thresholded ``pass'' counts is secondary to the scalar reward itself. In particular, the $R{=}1$ vs.\ $R{=}2$ story is less stable than on SkillsBench: the second iteration is clearly beneficial on \texttt{gru}, mildly beneficial on \texttt{deepnarrowmlp}, and weaker than $R{=}1$ on \texttt{shufflenet}. The correct takeaway is therefore narrower but still useful: \model's authoring loop is not restricted to binary workflow tasks, and can transfer into continuous-reward optimization domains when the reward is treated as a first-class signal.

\end{document}